\documentclass[sigconf, language=english,nonacm]{acmart}

\AtBeginDocument{%
  }

\usepackage{siunitx}
\usepackage{glossaries}
\usepackage{subcaption}
\usepackage{multirow}

\begin{document}

\renewcommand{\sectionautorefname}{Section}
\renewcommand{\subsectionautorefname}{Subsection}
\microtypesetup{expansion=true,protrusion=true}
\title[Knowledge Distillation for Efficient Transformer-Based Reinforcement Learning in Hardware-Constrained EMS]{Knowledge Distillation for Efficient Transformer-Based Reinforcement Learning in Hardware-Constrained Energy Management Systems}

\author{Pascal Henrich}
\email{pascal.henrich@student.kit.edu}
\orcid{0009-0004-3228-5685}
\affiliation{%
  \institution{Karlsruhe Institute of Technology}
  \city{Karlsruhe}
  \country{Germany}
}
\authornote{Corresponding authors}

\author{Jonas Sievers}
\email{jonas.sievers@kit.edu}
\orcid{0000-0003-2995-8602}
\affiliation{%
  \institution{Karlsruhe Institute of Technology}
  \city{Karlsruhe}
  \country{Germany}
}
\authornotemark[1]

\author{Maximilian Beichter}
\email{maximilian.beichter@kit.edu}
\orcid{0009-0009-9713-3590}
\affiliation{%
  \institution{Karlsruhe Institute of Technology}
  \city{Karlsruhe}
  \country{Germany}
}

\author{Thomas Blank}
\email{thomas.blank@kit.edu}
\orcid{0000-0002-7543-5653}
\affiliation{%
  \institution{Karlsruhe Institute of Technology}
  \city{Karlsruhe}
  \country{Germany}
}

\author{Ralf Mikut}
\email{ralf.mikut@kit.edu}
\orcid{0000-0001-9100-5496}
\affiliation{%
  \institution{Karlsruhe Institute of Technology}
  \city{Karlsruhe}
  \country{Germany}
}

\author{Veit Hagenmeyer}
\email{veit.hagenmeyer@kit.edu}
\orcid{0000-0002-3572-9083}
\affiliation{%
  \institution{Karlsruhe Institute of Technology}
  \city{Karlsruhe}
  \country{Germany}
}

\renewcommand{\shortauthors}{Henrich et al.}

\begin{abstract}
Transformer-based reinforcement learning has emerged as a strong candidate for sequential control in residential energy management. In particular, the Decision Transformer can learn effective battery dispatch policies from historical data, thereby increasing photovoltaic self-consumption and reducing electricity costs. However, transformer models are typically too computationally demanding for deployment on resource-constrained residential controllers, where memory and latency constraints are critical. The present paper investigates Knowledge Distillation to transfer the decision-making behaviour of high-capacity Decision Transformer policies to compact models more suitable for embedded deployment. Using the Ausgrid dataset, we train teacher models in an offline, sequence-based Decision Transformer framework on heterogeneous multi-building data. Given the trained models, we distil smaller student models by matching the teachers’ actions, thereby preserving control quality while reducing model size. Across a broad set of teacher–student configurations, distillation largely preserves control performance and even yields small improvements of up to \SI{1}{\percent}, while reducing the parameter count by up to \SI{96}{\percent}, the inference memory by up to \SI{90}{\percent}, and the inference time by up to \SI{63}{\percent}. Beyond these compression effects, comparable cost improvements are also observed when distilling into a student of identical architectural capacity. Overall, our results show that Knowledge Distillation makes Decision Transformer control more applicable for residential energy management on resource-limited hardware.
\end{abstract}

\newacronym{iai}{IAI}{Institute for Automation and Applied Informatics}
\newacronym{ipe}{IPE}{Institute for Data Processing and Electronics}

\newacronym{nsw}{NSW}{New South Wales}

\newacronym{bess}{BESS}{Battery Energy Storage System}
\newacronym{ems}{EMS}{Energy Management System}
\newacronym{hems}{HEMS}{Home Energy Management System}
\newacronym{bem}{BEMS}{Bulding Energy Management System}
\newacronym{hem}{HEM}{Home Energy Management}
\newacronym{ies}{IES}{Integrated Energy System}
\newacronym{vpp}{VPP}{Virtual Power Plant}
\newacronym{ev}{EV}{Electric Vehicle}
\newacronym{fcev}{FCEV}{Fuel Cell Electric Vehicle}
\newacronym{pv}{PV}{Photovoltaic}
\newacronym{res}{RES}{Renewable Energy Sources}
\newacronym{dsm}{DSM}{Demand-Side Management}
\newacronym{mpc}{MPC}{Model Predictive Control}

\newacronym{ml}{ML}{Machine Learning}
\newacronym{rl}{RL}{Reinforcement Learning}
\newacronym{td}{TD}{Temporal-Difference}
\newacronym{dp}{DP}{Dynamic Programming}
\newacronym{marl}{MARL}{Multi-Agent Reinforcement Learning}
\newacronym{dt}{DT}{Decision Transformer}
\newacronym{ddpg}{DDPG}{Deep Deterministic Policy Gradient}
\newacronym{sac}{SAC}{Soft Actor–Critic}
\newacronym{ppo}{PPO}{Proximal Policy Optimisation}
\newacronym{kd}{KD}{Knowledge Distillation}
\newacronym{llm}{LLM}{Large Language Models}
\newacronym{mdp}{MDP}{Markov Decision Process}
\newacronym{pomdp}{POMDP}{Partially Observable Markov Decision Process}
\newacronym{rnn}{RNN}{Recurrent Neural Network}
\newacronym{mlp}{MLP}{Multilayer Perceptron}
\newacronym{lstm}{LSTM}{Long Short-Term Memory}
\newacronym{gnn}{GNN}{Graph Neural Network}
\newacronym{milp}{MILP}{Mixed Integer Linear Programming}
\newacronym{gpt}{GPT}{Generative Pretrained Transformer}

\newacronym{soe}{SoE}{State of Energy}
\newacronym{ctg}{CtG}{Cost-to-Go}
\newacronym{rtg}{RtG}{Return-to-Go}

\begin{teaserfigure}
  \centering
  \includegraphics[width=\textwidth]{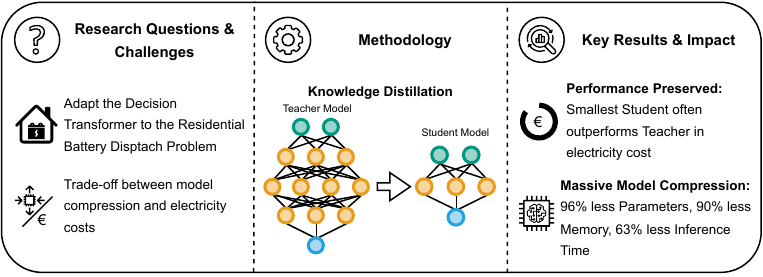}
  \caption{Graphical Abstract}
  \label{fig:teaser}
\end{teaserfigure}

\received{29 January 2026}
\received[revised]{XX XX 2026}
\received[accepted]{XX XX 2026}

\maketitle

\section{Introduction}
The increasing deployment of renewable energy sources, particularly \gls{pv} systems, introduces significant variability and temporal imbalances between electricity generation and household demand. To address these fluctuations, \glspl{hems} have become critical for coordinating flexible resources, such as \glspl{bess}, to lower operational costs and improve grid stability \cite{lund2015}. While traditional methods such as rule-based control \cite{jafari2020, abbasi2023} or \gls{mpc} \cite{yousefi2021,tammaru2024} are often used, they frequently struggle with highly variable conditions or rely on system models that are difficult to maintain in real-world settings \cite{zhang2022}.
As a data-driven alternative, \gls{rl} has gained prominence, with the \gls{dt} \cite{chen2021} emerging as a novel architecture that frames decision-making as a conditional sequence-modelling problem. By utilising a Transformer-based decoder to predict actions from past observations and target returns, the \gls{dt} provides a stable, goal-conditioned framework. However, the performance potential of Transformer architectures is fundamentally tied to their considerable model sizes, which can range from hundreds of millions to several billions of parameters, and to the amount of training data \cite{li2023}. Empirical evidence further indicates that increasing model capacity typically leads to improved performance \cite{lee2022}. At the same time, the resulting computational and memory demands pose a significant challenge for deployment on resource-constrained embedded hardware used in residential energy controllers.

To bridge the gap between architectural complexity and practical deployment, \gls{kd}, introduced as a formal term by \citet{hinton2015}, serves as a primary strategy for model compression. \gls{kd} employs a teacher-student paradigm, where a high-capacity teacher model transfers its learned behavioural patterns to a smaller, more efficient student. This process allows the student to mimic the teacher's internal representations, enabling the creation of compact models that preserve predictive control without the massive hardware requirements of the original architecture \cite{gou2021}. Building on this principle, this work explores the application of \gls{kd} to \glspl{dt} trained on aggregated multi-building datasets to leverage the needed diverse structural patterns and enhance policy robustness.

\subsection{Related Work}
Advances in computer vision \cite{chen2022,yang2024} and natural language processing \cite{sanh2020,xu2024} have established \gls{kd} as a key technique for transferring the capabilities of large, high-performing models into compact architectures. Building on these successes, recent research has increasingly explored the application of \gls{kd} in energy systems, where similar efficiency–performance trade-offs arise in edge-deployed controllers and forecasting pipelines. Accordingly, this section reviews recent applications of \gls{kd} across load identification, energy forecasting, battery health management, and distributed and \gls{rl}-based energy systems. Furthermore, we explore the domain of \glspl{dt}, review two papers combining them with \gls{kd}, and conclude with a discussion of the role of \glspl{dt} in energy systems.

\paragraph{Knowledge Distillation in Energy}In the domain of non-intrusive load monitoring, \gls{kd} is commonly employed to transfer knowledge from high-accuracy, cloud-based teacher models to lightweight student models deployed on smart meters: \citet{yi2025} propose a one-dimensional CNN–BiLSTM–at\-tention teacher model to guide the training of a compact student architecture, achieving a balance between identification accuracy and local computational constraints. To further reduce model complexity, \citet{zhao2025} introduce an integrated pruning and distillation framework, IPD-NILM, which combines layer-wise iterative Fisher pruning with distillation and reduces the number of trainable parameters by \SI{72}{\percent} while maintaining comparable performance. Moving beyond compression alone, \citet{batic2025} incorporate perception-aligned gradients into the distillation process to ensure that student models inherit reliable and human-interpretable decision patterns, addressing the opacity of deep neural networks deployed at the edge. Beyond load identification, \gls{kd} has also been applied to energy forecasting and battery health management in order to enable the local deployment of complex time-series and graph-based models. \citet{chan2025} demonstrate the distillation of large Transformer-based foundation models into compact expert models for battery capacity degradation forecasting, enabling cross-capacity generalisation suitable for onboard battery management systems. Similarly, \citet{lin2025} distil spatial–temporal knowledge from graph neural networks pretrained on utility-scale datasets into household-level multilayer perceptrons, alleviating data scarcity at the local level. In the context of dynamic multi-energy microgrids, \citet{bao2024} employ KD-LSTM architectures to adapt to evolving operating conditions, extracting salient features from sparse data while significantly reducing computational resource consumption. In distributed learning environments, particularly federated learning, \gls{kd} has been shown to mitigate challenges arising from non-independent and identically distributed data and heterogeneous hardware capabilities. In the context of short-term forecasting in electric Internet-of-Things systems, \citet{tong2025} propose hierarchical \gls{kd} to identify and distil knowledge from a subset of top-performing clients that best represent the global gradient direction, improving convergence stability in electric Internet-of-Things systems. For energy theft detection in smart grids, \citet{zou2023} introduce an edge-assisted federated contrastive distillation approach in which local teacher models guide student training, reducing detection loss without requiring centralised data aggregation. \gls{kd} has also been explored in \gls{rl}–based energy management to enhance policy interpretability. In the domain of building energy management, \citet{chen2025} propose value-aware distillation to extract interpretable decision rules from complex Soft Actor-Critic policies into decision tree ensembles, preserving over \SI{91}{\percent} of the teacher’s performance while improving transparency for system operators.

\paragraph{Knowledge Distillation and Decision Transformer} In the domain of offline multi-agent \gls{rl}, \citet{tseng2022} train a centralised \gls{dt} teacher with access to all agents’ observations, actions, and rewards, and subsequently distil both action predictions and inter-agent relational structure into decentralised student policies, yielding improved convergence and robustness over sequence-modelling and offline RL baselines. In the context of visual robotic manipulation, \citet{chen2024a} propose a transformer-based \gls{kd} framework in which a state-based decision-transformer teacher transfers control policies to an image-based student via state estimation and weight initialisation, enabling effective policy learning under partial observability.

\paragraph{Decision Transformer in Energy Management} More recently, \glspl{dt} have gained traction in energy management by reframing sequential control as a conditional sequence modelling problem, for example, in integrated energy systems \cite{li2024} and building-level energy management \cite{zhang2025}. 

\subsection{Contributions}
In light of these findings, while \gls{kd} has proven effective across a range of energy applications, its systematic integration with \glspl{dt} for residential battery dispatch remains largely unexplored. Existing DT-based approaches in energy systems primarily focus on architectural design choices related to model size rather than on compression methods such as \gls{kd}. We address this gap by demonstrating that \gls{kd} can substantially reduce the computational overhead of \glspl{dt} while preserving control performance, thereby enabling their deployment on resource-constrained energy management hardware. Our main contributions are:

\begin{itemize}
    \item We adapt the \gls{dt} architecture to the residential battery dispatch problem and analyse the impact of model size on control performance, showing that \textit{medium}-sized models yield the strongest cost reductions.
    \item We propose a response-based \gls{kd} framework for compressing \glspl{dt} and demonstrate reductions of up to \SI{96}{\percent} in parameters, \SI{90}{\percent} in inference memory, and \SI{63}{\percent} in inference time, while largely preserving control performance and even improving it by up to \SI{1}{\percent}.
    \item We investigate offline self-distillation and show that it acts as an effective regulariser. 
\end{itemize} 

All results are benchmarked against a state-of-the-art \gls{rl} baseline based on \gls{ddpg}, a theoretical lower bound (ideal case with perfect foresight) via \gls{milp}, and an upper bound established by a standard rule-based controller. The remainder of the present paper is organised as follows: \autoref{sec:method} details the underlying method of the \gls{dt} and the distillation process; \autoref{sec:experiments} describes the experiments with results; \autoref{sec:discussion} provides a discussion of the findings and their implications; and \autoref{sec:conclusion} presents the conclusions.

\section{Method}
\label{sec:method}
This section presents the fundamentals of the proposed approach to residential battery scheduling based on \gls{dt} and model compression, designed to enable deployment in \glspl{hems} operating under computational and memory constraints.

\subsection{Reinforcement Learning with Decision Transformers}
\begin{figure}[t!]
    \centering
    \includegraphics[width=\linewidth]{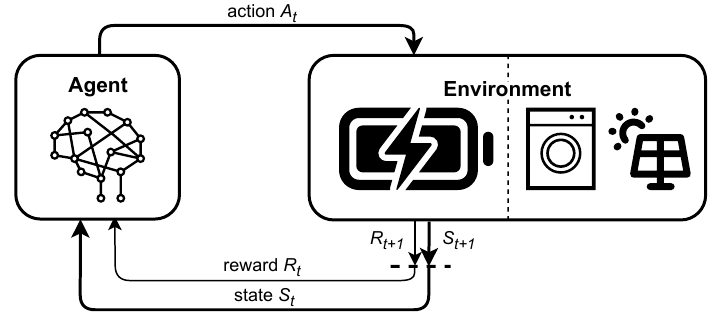}
    \caption{Markov Decision Process for Reinforcement Learning in a Battery Scheduling Environment.}
    \label{fig:mdp}
\end{figure}

\gls{rl} is a control and decision-making paradigm in which an agent interacts sequentially with an environment by observing system states $S_t \in \mathcal{S}$, selecting actions $A_t \in \mathcal{A}$ according to a policy $\pi$, and receiving scalar feedback in the form of rewards $R_t \in \mathcal{R}$. This interaction is commonly formalised as a \gls{mdp} (\autoref{fig:mdp}), defined by the tuple $(\mathcal{S}, \mathcal{A}, p, \mathcal{R}, \gamma)$, where $p$ denotes the state transition dynamics and $\gamma \in [0,1)$ is a discount factor. The objective of \gls{rl} is to learn a policy $\pi^\ast$ that maximises the expected cumulative discounted reward.

A central concept in \gls{rl} is the state-value function, which evaluates the expected return when following a policy $\pi$ from a given state $s$:
\begin{equation}
    V^{\pi}(s) = \mathbb{E}_{\pi} \left[ \sum_{k=0}^{\infty} \gamma^{k} R_{t+k} \,\middle|\, S_t = s \right].
\end{equation}
In problems with large or continuous state spaces, the value function is commonly approximated using a parameterised function $V(s;\theta)$, referred to as value function approximation.

\gls{td} learning is a widely used approach for estimating value functions that updates predictions incrementally based on experience. Rather than waiting until the end of an episode to observe complete returns, \gls{td} methods adjust value estimates at each time step by combining the immediately observed reward with the current estimate of the value of the subsequent state. This reliance on bootstrapping enables efficient online learning. However, because \gls{td} methods update predictions using other learned predictions, approximation errors can propagate over time.

In the context of residential energy management, \gls{rl} enables adaptive control strategies for \glspl{hems} by learning directly from operational data how to coordinate flexible resources in response to household demand, \gls{pv} generation, and time-varying electricity prices. By encoding electricity costs within the reward signal, the learned policy naturally aims to minimise long-term electricity costs while respecting the underlying system dynamics.

This form of sequential decision-making is commonly referred to as online \gls{rl}, in which an agent collects data through direct interaction with the environment during training. Within this setting, a further distinction is made between on-policy and off-policy methods, which differ in whether the training data is generated by the current policy or by a separate behaviour policy. However, in many real-world applications, including residential energy management, generating data on the fly is impractical or undesirable due to safety, cost, or operational constraints. This motivates a shift towards offline \gls{rl}, in which policies are learned exclusively from a fixed dataset collected prior to training, without further interaction with the environment.

Traditional \gls{rl} methods commonly employ value function approximation alongside \gls{td} learning \cite{mnih2013, lillicrap2019}. When combined with off-policy data, this setting can lead to instability arising from the interaction of bootstrapping, approximation errors, and distribution mismatch—an issue widely known as the deadly triad \cite{sutton1998}. In offline \gls{rl}, where learning is performed solely from fixed datasets without online interaction, these effects are exacerbated, as the absence of environment feedback increases sensitivity to extrapolation errors \cite{levine2020}.
\begin{figure}[t!]
    \centering
    \includegraphics[width=\linewidth]{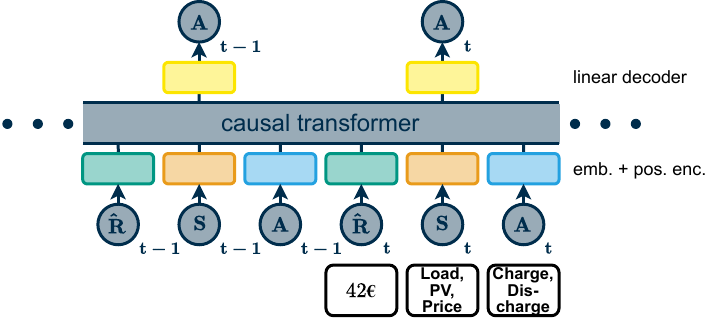}
    \caption{Decision Transformer in a Battery Scheduling Environment (adapted from \cite{chen2021}).}
    \label{fig:dt}
\end{figure}

To mitigate these issues, we adopt the \gls{dt} (\autoref{fig:dt}) proposed by \citet{chen2021}, which reformulates \gls{rl} as a conditional sequence modelling problem. Rather than estimating value functions, the \gls{dt} uses a causal, decoder-only Transformer architecture \cite{vaswani2017} to directly predict actions from sequences of past states, actions, and returns. The Transformer is based on self-attention mechanisms, which enable the model to capture long-range temporal dependencies across entire trajectories more effectively than recurrent architectures. The self-attention mechanism allows each timestep to selectively attend to all previous elements in the sequence, learning relevance weights. To ensure an autoregressive policy structure, the \gls{dt} applies a causal attention mask that prevents information leakage from future timesteps. This causal constraint is essential for sequential decision-making, as it enforces that action predictions at time step $t$ depend only on information available up to that point. As a result, the model remains consistent with the underlying \gls{mdp} formulation while benefiting from the expressive power of Transformer's sequence modelling. 

This supervised learning formulation avoids explicit bootstrapping and reduces dependence on off-policy value estimation, thereby improving stability in offline \gls{rl} settings. A core innovation of the \gls{dt} is its use of future-oriented conditioning to guide decision-making. Rather than relying on retrospective reward signals and \gls{td} updates as in conventional \gls{rl}, the \gls{dt} directly conditions the policy on a desired measure of future performance. In its original formulation from \citet{chen2021}, this is achieved through the \gls{rtg}, which at time step $t$ is defined as the cumulative sum of all future rewards:
\begin{equation}
\label{eq:rtg}
\hat R_t = \sum_{t' = t}^{T} R_{t'} .
\end{equation}
By conditioning action predictions on $\hat R_t$, the \gls{dt} learns to generate action sequences that are explicitly directed toward achieving a specified return. As a result, discounting is not required to ensure convergence. Instead, the Transformer architecture enables direct credit assignment via self-attention, allowing the model to capture long-range dependencies between states, actions, and returns across the entire trajectory.

Under this formulation, a trajectory $\tau$ is represented as an alternating sequence of \gls{rtg}, state, and action tuples,
\begin{equation}
\label{eq:traj_rtg}
\tau = (\hat R_1, S_1, A_1, \hat R_2, S_2, A_2, \dots, \hat R_T, S_T, A_T),
\end{equation}
enabling the model to learn the temporal dependencies between targeted return, observed system states, and the corresponding control actions required for cost-efficient battery operation.

\subsection{Knowledge Distillation for Model Compression}
\begin{figure}[t!]
    \centering
    \includegraphics[width=\linewidth]{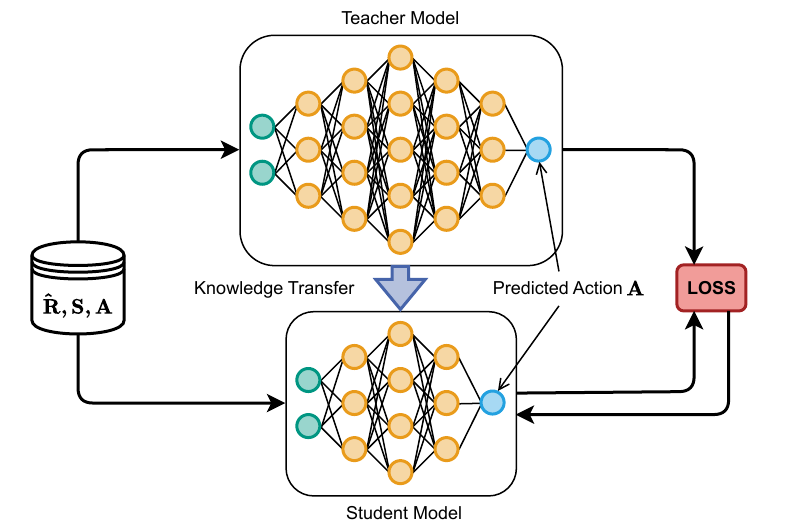}
    \caption{Response-Based Knowledge Distillation for Decision Transformer in a Battery Scheduling Environment.}
    \label{fig:kd}
\end{figure}

To enable deployment on resource-constrained embedded hardware, we apply \gls{kd} to transfer the behaviour of high-capacity \gls{dt} teacher models to compact student models. The primary focus of the present work is response-based \gls{kd} (\autoref{fig:kd}), in which the student is trained to replicate the outputs of a pretrained teacher model:
\begin{equation}
L_{\mathrm{Res}}(z_t, z_s) = L(z_t, z_s),
\end{equation}
where $z_t$ and $z_s$ denote the teacher and student logits (output of the neural networks), respectively, and $L(\cdot,\cdot)$ denotes the loss function used to match the outputs \cite{gou2021}.

Most popular response-based distillation approaches rely on temperature-scaled soft targets \cite{gou2021}. Temperature scaling is commonly used in classification-based \gls{kd} to soften the teacher’s output distribution by dividing the logits by a temperature parameter, thereby revealing relative class confidences and inter-class similarities that provide richer training signals to the student \cite{hinton2015}. However, this mechanism is inherently tied to discrete probability distributions produced by a softmax operation and does not naturally extend to regression-based control problems. Therefore, such techniques are not applicable in our setting, as the battery scheduling task involves continuous-valued control actions as logits rather than categorical class probabilities. Consequently, the student model is trained to directly match the teacher’s continuous outputs. Distillation is conducted in an offline setting, with the teacher model kept fixed throughout student training.

In addition to conventional teacher–student distillation, we also consider a special case, self-distillation. In this setting, the student model is trained to mimic the outputs of a pretrained teacher with an identical architectural configuration. Unlike other knowledge distillation approaches, this procedure does not reduce the model size; instead, the offline self-distillation process acts as a form of regularisation, encouraging smoother, more consistent internal representations and thereby improving generalisation performance without increasing model capacity or computational complexity \cite{furlanello2018}.

\section{Evaluation}
\label{sec:experiments}
This section presents the experimental evaluation of \gls{kd} applied to \gls{dt} as a sequence-modelling-based approach for \gls{bess} control. The evaluation assesses the \gls{dt}'s ability to exceed its behavioural baseline and quantifies the effectiveness of \gls{kd} in facilitating deployment on resource-constrained hardware.

\subsection{Problem Formulation -- Environment}
\label{subsec:problem_formulation}
The operation of a residential \gls{bess} is formulated as a finite-horizon \gls{mdp} to enable the application of \gls{rl}. This formulation models the sequential decision-making process of an autonomous control agent interacting with a residential energy system, aiming to minimise cumulative electricity-buying costs over the planning horizon.

The environment dynamics are derived from high-resolution time-series data, including household electricity consumption, \gls{pv} generation, and wholesale electricity prices. Prosumption is defined as the net electrical power exchange between the household and the grid, where positive values indicate grid imports and negative values represent surplus \gls{pv} generation. Battery energy capacities are defined on a per-building basis according to the mean daily surplus \gls{pv} energy. The corresponding nominal power ratings are then derived by constraining the maximum charging and discharging power to one-quarter of the battery energy capacity.

At each timestep $t$, the agent observes a continuous-valued state vector $S_t$ that captures the current physical status of the system together with limited foresight provided by forecasts. The state is defined as
\begin{equation}
\label{eq:states_vector}
\begin{aligned}
S_t = (&SoE_t, Prosumption_t, \dots, Prosumption_{t+1+h}, \\
       & Price_t, \dots, Price_{t+1+h})
\end{aligned}
\end{equation}
where $SoE_t$ denotes the battery \gls{soe} and $h$ represents the forecast horizon. In this work, $h$ is set to 24 time slots, corresponding to a 12-hour look-ahead, to support informed decision-making.

The action $A_t \in \mathcal{A}$ is a continuous variable representing the intended power flow to or from the battery. Positive values correspond to charging, while negative values indicate discharging. Actions are constrained to the interval $[-Power_{max}/2,\, Power_{max}/2]$, where the division by 2 accounts for the conversion from power in \SI{}{\kilo\watt} to energy exchanged over a 30-minute control interval. To ensure physical feasibility, the environment enforces deterministic battery dynamics that clip actions violating \gls{soe} constraints. The resulting effective action $\tilde{A}_t$ is defined as
\begin{equation}
\tilde{A}_t = \text{clip}(SoE_t + A_t,\, 0,\, Capacity_{max}) - SoE_t.
\end{equation}
The resulting grid exchange $g_t$ is computed as the sum of net prosumption and the effective battery action: $g_t = Prosumption_t + \tilde{A}_t$.

The reward function is economically motivated and aims to minimise the net monetary cost $c_t$. It is defined as
\begin{equation}
R_t = -c_t - \|A_t - \tilde{A}_t\|_1.
\end{equation}
Grid imports ($g_t \geq 0$) are charged at the time-varying wholesale electricity price, while grid exports ($g_t < 0$) are recompensed at a fixed feed-in tariff of \SI{0.1}{\text{€}\per\kilo\watt\hour}. The penalty term $\|A_t - \tilde{A}_t\|_1$ discourages the selection of actions that cannot be physically implemented due to battery constraints.

When utilising the \gls{dt}, the standard \gls{mdp} trajectory formulation is reframed as a conditional sequence-modelling problem. Rather than conditioning on the conventional \gls{rtg} (\autoref{eq:rtg}), the model is conditioned on the \gls{ctg}, denoted by $\hat{C}_t$, which is defined as the cumulative sum of future costs
\begin{equation}
\hat{C}_t = \sum_{t'=t}^{T} c_{t'}.
\end{equation}
Under this formulation, the trajectory $\tau$ (\autoref{eq:traj_rtg}) is represented as
\begin{equation}
    \tau = (\hat{C}_1, S_1, A_1, \dots, \hat{C}_T, S_T, A_T),
\end{equation}
enabling the model to autoregressively generate actions that correspond to specified future cost targets. With respect to the network design, rather than following \citet{chen2021} in employing a GPT-2–based model \cite{radford2019} with its adapted architecture and Byte Pair Encoding, we adopt the original Transformer architecture proposed by \citet{vaswani2017}. Nevertheless, we retain the embedding scheme and positional encoding strategy introduced by \citet{chen2021} to preserve compatibility with trajectory-based sequence modelling.

We formulate \gls{kd} as a supervised policy-matching problem in which a compact student policy $\pi_S$ is trained to approximate the behaviour of a high-capacity teacher policy $\pi_T$. Both policies are induced by their respective \gls{dt} models and operate over the same state space, aiming to transfer decision-making competence while reducing model complexity.

Given an offline dataset $\mathcal{D}$ of trajectories, the distillation objective enforces alignment between the student and teacher policies at the level of their action-generating logits. Specifically, the student policy $\pi_S$ is said to converge to the teacher policy $\pi_T$ under the data distribution $\mathcal{D}$ if the student parameters $\theta_S^\star$ minimise the expected discrepancy between the corresponding model outputs, as defined by

\begin{equation}
\label{eq:kd_form}
    \pi_S \overset{\mathcal{D}}{\longrightarrow} \pi_T
    \Longleftrightarrow
    \theta_S^\star \in \arg\min_{\theta_S}
    \mathbb{E}_{S_t \sim \mathcal{D}}
    \left[
    \operatorname{SL1}\left(z_S(S_t)- z_T(S_t)\right)
    \right],
\end{equation}
where $z_S(S_t)$ and $z_T(S_t)$ denote the scalar action logits produced by the student and teacher models for a given state $S_t$. The Smooth L1 loss $\operatorname{SL1}(\cdot)$ \cite{girshick2015} is employed to provide robustness to outliers while preserving sensitivity to small deviations, which is particularly beneficial in continuous-action settings \cite{maggipinto2020}. By optimising this objective, the student is encouraged to replicate the teacher’s policy behaviour over the empirical state distribution induced by $\mathcal{D}$.

\subsection{Data and Hardware Setup}
To ensure the reproducibility of our results, we describe the underlying data sources, the methodology for generating the offline dataset, and the hardware infrastructure used for the experiments.

\begin{figure}[t!]
    \centering
    \includegraphics[width=\linewidth]{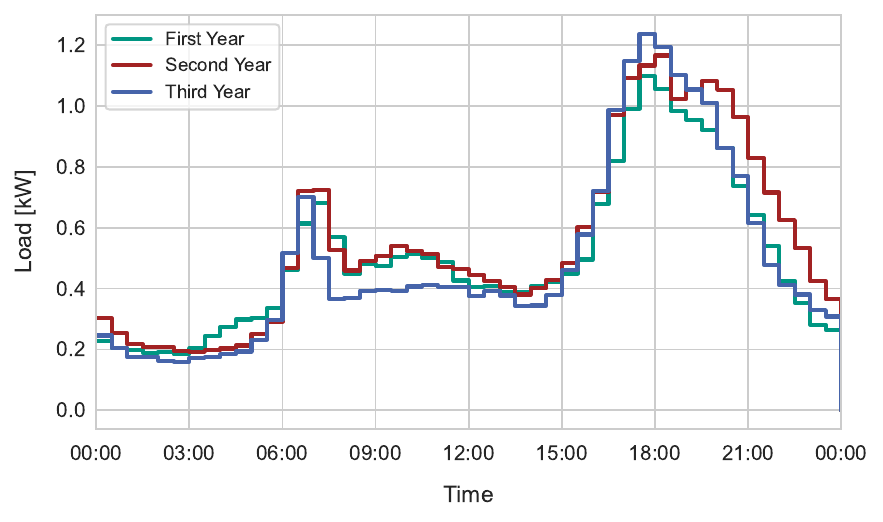}
    \caption{Aggregated Daily Mean Load for Building 13 with a Typical Residential Load Pattern.}
    \label{fig:ausgrid_load}
\end{figure}

\paragraph{Used Dataset} All of the following experiments are conducted using fixed random seeds (42, 1894, and 314159) to ensure reproducibility and statistical robustness, using data from 20 residential buildings (building numbers: 13, 20, 33, 35, 74, 75, 82, 87, 88, 101, 106, 109, 130, 144, 152, 153, 157, 161, 169, 176) selected from the Ausgrid Solar Home Electricity Dataset \cite{ratnam2017} in combination with wholesale electricity prices of DE/AT/LU obtained from the SMARD database \cite{smard}. \autoref{fig:ausgrid_load} illustrates the load profile of Building 13 as a representative example. 

\paragraph{Offline Dataset Generation} As the \gls{dt} operates in an offline learning setting, it relies on a pre-collected behavioural dataset $\mathcal{D}$. To generate this dataset under operating conditions, a \gls{ddpg} agent is employed as the trajectory-generating policy. The \gls{ddpg} agent interacts with the environment defined in \autoref{subsec:problem_formulation}, producing state--action trajectories that capture the underlying system dynamics and operational constraints. To facilitate comparison with established benchmarks, we adopt the \gls{ddpg} hyperparameter configuration reported in \cite{sievers2025}, whereas the most critical \gls{dt} and \gls{kd} hyperparameters can be found in \autoref{sec:hyperparameter}.

\paragraph{Used Hardware} The computational workload is distributed across two specialised compute nodes. Trajectory generation using \gls{ddpg} is performed on a CPU node equipped with an AMD EPYC~9454 processor featuring 96 cores and \SI{384}{\gibi\byte} of main memory. All Transformer-based experiments, including \gls{kd}, are executed on a GPU node composing an Intel Xeon Platinum~8358 processor with 64 cores, \SI{512}{\gibi\byte} of system memory, and four NVIDIA H100 accelerators, each providing \SI{94}{\gibi\byte} of device memory.

\subsection{Experiments}
\begin{table}[t!]
\centering
\caption{Hyperparameter of Different Decision Transformer Model Sizes.}
\label{tab:dt_parameter}
\begin{tabular}{@{}llll@{}}
\toprule
\textbf{Name} & \textbf{\# Layers} & \textbf{\# Heads} & \textbf{Model Dimension} \\ \midrule
\textit{Tiny} & 1 & 1 & 64  \\
\textit{Mini} & 2 & 2 & 128 \\
\textit{Small} & 3 & 1 & 128 \\
\textit{Medium} & 8 & 2 & 256 \\
\textit{Large} & 12 & 4 & 512 \\ \bottomrule
\end{tabular}
\end{table}
The experimental evaluation is conducted over a four-week period and is designed to assess the performance of the \gls{dt} across multiple model sizes, evaluate the effectiveness of \gls{kd} for model compression, and compare the final results against a benchmark.

\paragraph{Decision Transformer.}
To analyse the relationship between model capacity and control performance, as well as the necessity of \gls{kd}, five \gls{dt} variants with varying architectural depths and widths are considered. The corresponding architectural configurations are summarised in \autoref{tab:dt_parameter}.

\paragraph{Knowledge Distillation.} 
The \gls{kd} experiments follow an offline distillation paradigm in which high-capacity teacher models are first trained to convergence before transferring their knowledge to more compact student models. We adopt a response-based distillation strategy, where student models are trained to match the teacher's continuous action logits using a Smooth L1 loss, as formalised in \autoref{eq:kd_form}. Distillation is performed using three teacher configurations of increasing capacity: \textit{small}, \textit{medium}, and \textit{large}. Each teacher is distilled into three student architectures of reduced complexity, namely \textit{small}, \textit{mini}, and \textit{tiny}. This experimental design enables a systematic evaluation of how teacher model capacity influences the quality of the transferred supervision signal and the extent to which student models can retain control performance under compression. In addition, a specialised form of offline self-distillation is considered by distilling a pre-trained \textit{small} teacher into a student model of identical architecture. This setting is used to investigate the regularisation benefits of distillation in the absence of explicit model compression.

\subsection{Results}
This section presents the comparative evaluation of the proposed \gls{dt} and its knowledge-distilled variant against two benchmark control approaches. We first report the aggregate performance comparison and subsequently analyse the experimental results that lead to these outcomes, beginning with the standalone \gls{dt} and followed by the \gls{kd} experiments.

\begin{figure}[t!]
    \centering
    \includegraphics[width=\linewidth]{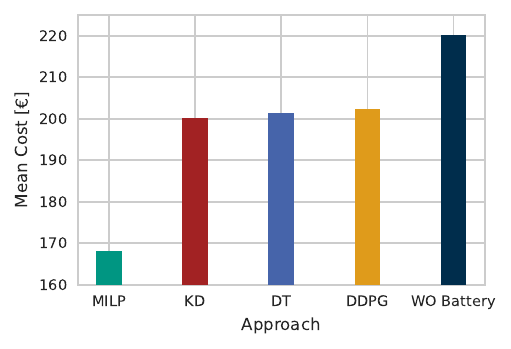}
    \caption{Mean Cost Across All 20 Buildings for Different Approaches. DT Denotes the \textit{Medium} Decision Transformer, while KD Refers to Knowledge Distillation From \textit{Medium} Teacher to \textit{Small} Student.}
    \label{fig:comp_best}
\end{figure}
\paragraph{Decision Transformer.} \autoref{fig:comp_best} summarises the results of all experiments by reporting the mean electricity cost over a four-week evaluation period, averaged across all buildings. All three learning-based approaches achieve performance between the two benchmark baselines, namely the without-battery scenario with a mean cost of \SI{220.25}{\text{€}} and the optimisation-based \gls{milp} (\autoref{sec:milp}) benchmark with a cost of \SI{168.08}{\text{€}}. Among the learning-based methods, the \gls{ddpg}-generated trajectory yields the weakest result, with a mean cost of \SI{202.43}{\text{€}}. The \gls{dt} improves upon this result with a mean cost of \SI{201.30}{\text{€}}, while the knowledge-distilled model achieves the best performance among the considered approaches, attaining a mean cost of \SI{200.28}{\text{€}}.

To contextualise these aggregate results, we next examine the individual experiments that lead to this performance, beginning with an analysis of the \gls{dt} models and subsequently turning to the \gls{kd} experiments. 

\begin{figure}[t!]
    \centering
    \includegraphics[width=\linewidth]{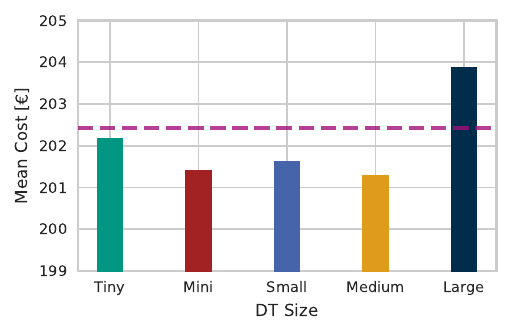}
    \caption{Mean Cost Across All 20 Buildings for Different Decision Transformer Sizes.}
    \label{fig:dt_mean_cost_sizes}
\end{figure}
As a first step in this analysis, we evaluate five \gls{dt} model variants with increasing architectural capacity to provide a suitable baseline for the subsequent distillation experiments. \autoref{fig:dt_mean_cost_sizes} illustrates the relationship between model size and mean cost performance across the evaluated \gls{dt} architectures, with the corresponding \gls{ddpg}-based trajectory performance indicated by the dashed horizontal reference line. The results reveal a non-monotonic relationship between model capacity and cost, therefore suggesting that larger models do not necessarily yield improved control performance. Notably, the four smaller \gls{dt} models consistently outperform the \gls{ddpg} trajectory baseline, whereas the \textit{large} model exhibits a higher mean cost when averaged across all 20 buildings. Among all evaluated configurations, the \textit{medium} \gls{dt} achieves the lowest mean cost. Beyond its favourable performance characteristics, the \textit{medium}-sized \gls{dt} also represents a practical choice for subsequent \gls{kd} experiments, as it provides sufficient representational capacity to serve as an effective teacher while remaining large enough to be compressible. Accordingly, this configuration is selected as the primary model for \gls{kd} and used for the final comparative evaluation shown in \autoref{fig:comp_best}.

\begin{figure}[t!]
    \centering
    \includegraphics[width=\linewidth]{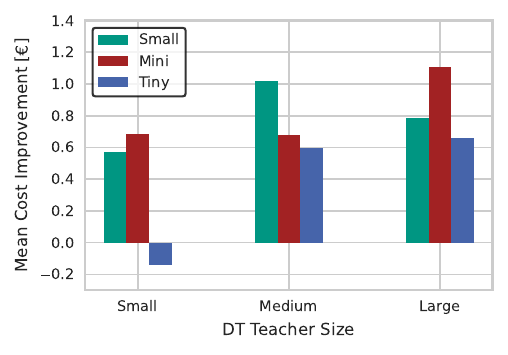}
    \caption{Mean Cost Improvement Across All 20 Buildings for Different Knowledge Distillation Sizes.}
    \label{fig:kd_mean_cost_improvement}
\end{figure}
\paragraph{Knowledge Distillation.} Building on the preceding analysis, we examine the three largest \gls{dt} teacher configurations, \textit{small}, \textit{medium}, and \textit{large}, and their ability to transfer knowledge to more compact student models. \autoref{fig:kd_mean_cost_improvement} reports the absolute cost improvements achieved by the distilled student models relative to their respective teachers.

Across the nine evaluated distillation configurations, eight result in improved student performance compared to the corresponding teacher. Distillation from the \textit{small} teacher yields the smallest overall improvements, with cost reductions of \SI{0.57}{\text{€}} for the \textit{small} student, \SI{0.69}{\text{€}} for the \textit{mini} student, and a slight cost increase of \SI{0.14}{\text{€}} for the \textit{tiny} student. The \textit{small}-to-\textit{small} configuration corresponds to a self-distillation setting and achieves a measurable performance gain over the identically sized teacher.

Distillation from the \textit{medium} teacher exhibits a monotonic decrease in improvement as student model size decreases. The \textit{small} student achieves the largest reduction in mean cost at \SI{1.02}{\text{€}}, followed by the \textit{mini} student with \SI{0.68}{\text{€}} and the \textit{tiny} student with \SI{0.60}{\text{€}}.

The \textit{large} teacher produces the highest overall performance gains among the evaluated configurations. Distillation into the \textit{small}, \textit{mini}, and \textit{tiny} student models reduces the mean cost by \SI{0.79}{\text{€}}, \SI{1.10}{\text{€}}, and \SI{0.66}{\text{€}}, respectively. When considered with the \gls{dt} performance reported in \autoref{fig:dt_mean_cost_sizes}, the distillation from the \textit{medium} teacher to the \textit{small} student yields the lowest absolute mean cost across all evaluated approaches.

\begin{figure}[t!]
    \centering
    \includegraphics[width=\linewidth]{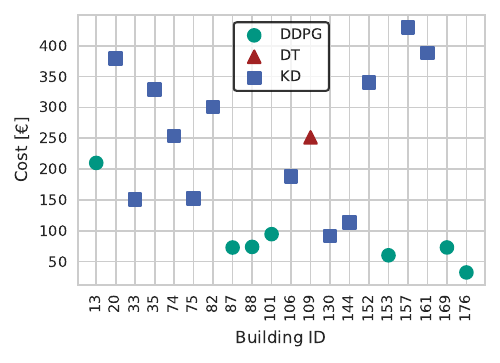}
    \caption{Best Performing Learning-Based Approach for Each of the 20 Buildings. DT Denotes the \textit{Medium} Decision Transformer, while KD Refers to Knowledge Distillation From \textit{Medium} Teacher to \textit{Small} Student.}
    \label{fig:comp_best_over_buildings_small}
\end{figure}
Based on the preceding results, the distillation from the \textit{medium} teacher to the \textit{small} student is identified as the most effective configuration and is henceforth referred to as the \gls{kd} model. A building-level comparison of all learning-based approaches is presented in \autoref{fig:comp_best_over_buildings_small}. The results indicate that the \gls{kd} model outperforms both \gls{ddpg} and \gls{dt} in \SI{60}{\percent} of the cases. The \gls{ddpg} model achieves the best performance for \SI{35}{\percent} of the buildings, while \gls{dt} is superior in only a single building.

\begin{table}[t!]
\centering
\caption{Model Size Comparison in Terms of Total Parameters, Memory Footprint, and Inference Time for Different Decision Transformer Variants.}
\label{tab:model_stats}
\begin{tabular}{@{}llll@{}}
\toprule
 & Total Parameters & Total Memory [\SI{}{\mega\byte}] & Time [ms] \\ \midrule
\textit{Tiny} & \num{387713} & \num{7.09} & \num{3.56}\\
\textit{Mini} & \num{1498241} & \num{17.44} & \num{4.80} \\
\textit{Small} & \num{2157569} & \num{25.45} & \num{5.99} \\
\textit{Medium} & \num{9831681} & \num{70.56} & \num{9.66} \\
\textit{Large} & \num{51166721} & \num{219.83} & \num{16.83} \\ \bottomrule
\end{tabular}
\end{table}
For the final evaluation regarding deployment on resource-con\-strained hardware, we analyse the relationships among model complexity, memory requirements, and inference latency. \autoref{tab:model_stats} reports the total number of parameters, the estimated memory footprint required for inference (computed using the \texttt{torchinfo} package \cite{torchinfo2023}), and the measured inference time of the considered models across the four-week evaluation period, averaged per timestep. The results demonstrate that the knowledge-distilled model reduces the parameter count by \SI{78}{\percent}, the inference-time memory footprint by \SI{64}{\percent}, and the inference latency by \SI{38}{\percent}, while simultaneously achieving lower electricity costs for the majority of evaluated buildings. While the \textit{medium}-to-\textit{small} configuration represents the overall best-performing model, even the distillation from the \textit{medium} teacher to the \textit{tiny} student (\autoref{fig:comp_best_over_buildings_tiny}), corresponding to the smallest evaluated architecture, achieves the lowest cost for \SI{55}{\percent} of the buildings. In this configuration, the parameter count is reduced by \SI{96}{\percent}, the inference-time memory footprint by \SI{90}{\percent}, and the inference latency by \SI{63}{\percent}.

\section{Discussion}
\label{sec:discussion}
This present paper introduces the theoretical concept of \gls{kd} and its practical implementation in a \gls{dt} for a residential battery storage setup. Among the studies conducted, we demonstrate that integrating \gls{dt} and \gls{kd} provides a robust framework for \gls{bess} control, effectively bridging the gap between the high-capacity architectural potential and the strict requirements of embedded hardware.

A primary finding of this work is that knowledge-distilled models do not merely approximate their teachers but consistently outperform them. While the \textit{medium} \gls{dt} achieved a mean cost of \SI{201.30}{\text{€}}, the \gls{kd} model (\textit{medium}-to-\textit{small}) reduced this to \SI{200.28}{\text{€}}, making it the top-performing learning-based approach. This performance boost, even observed in self-distillation settings where the student and teacher share identical architectures, suggests a significant regularisation effect. By training on the teacher's continuous action logits rather than the original dataset targets, the student model is likely to learn a smoother, more generalised decision. This allows the student to bypass specific noise or suboptimal patterns present in the \gls{ddpg}-generated training trajectories.

The analysis of different \gls{dt} sizes reveals a non-monotonic relationship between model capacity and control performance. Interestingly, the \textit{medium} \gls{dt} outperformed the \textit{large} \gls{dt}, which actually performed worse than the \gls{ddpg} baseline in aggregate. This suggests that for the specific task of residential battery scheduling, over-parameterisation may lead to overfitting on building-specific idiosyncrasies in the offline dataset, hindering the model's ability to generalise across heterogeneous residential environments. The \textit{medium} architecture appears to offer the optimal range of representational capacity for capturing structural regularities such as \gls{pv} patterns and seasonal effects.

The most critical implication for real-world deployment is the massive reduction in computational overhead achieved through \gls{kd}. The \textit{medium}-to-\textit{small} configuration provides a beneficial reduction in parameters, memory size and inference latency while achieving the best economic results. For even more restrictive hardware, the \textit{medium}-to-\textit{tiny} distillation offers an extreme compression case, where the parameter count is reduced by \SI{96}{\percent}, the inference-time memory footprint by \SI{90}{\percent}, and the inference latency by \SI{63}{\percent}. Despite this radical reduction in size, the \textit{tiny} student still outperformed benchmarks in \SI{55}{\percent} of the evaluated buildings. These results confirm that \gls{kd} effectively extracts the knowledge of large sequence models into compact variants that are viable for low-cost embedded controllers without sacrificing reliability.

While all learning-based approaches remained within the bounds set by the no-battery baseline and the theoretical \gls{milp} optimum, the \gls{kd} model proved to be the most robust. The fact that the \gls{kd} model outperformed the original \gls{ddpg} trajectories, which were used to train the \gls{dt} in the first place, highlights the \gls{dt}’s ability to frame \gls{rl} as a sequence-modelling problem to overcome the deadly triad of instability and extrapolation errors common in \gls{rl}. 

In conclusion, these findings demonstrate that \gls{kd} is not just a tool for efficiency, but a necessary refinement step that enhances the robustness and generalizability of Transformer-based controllers in dynamic energy environments.

\section{Conclusion}
\label{sec:conclusion}
In the present paper, we introduce a novel \acrfull{kd} framework to compress \acrfull{dt} policies for residential battery energy storage control under strict hardware constraints. Using offline trajectories from the Ausgrid dataset, we train high-capacity teacher models on aggregated multi-building data and distil compact student models by matching the teachers’ control actions. Our results show that \gls{kd} can preserve \gls{dt} control quality while drastically reducing model complexity. Across 20 buildings, the distilled policies achieve performance comparable to their teachers and even outperform them for up to \SI{60}{\percent} of the buildings. At the same time, distillation delivers substantial deployment benefits. The best performing \textit{medium}-to-\textit{small} student reduces the parameter count by \SI{78}{\percent}, the inference-time memory footprint by \SI{64}{\percent}, and the inference latency by \SI{38}{\percent} while maintaining strong control performance. Even under extreme compression, the \textit{medium}-to-\textit{tiny} student achieves reductions of \SI{96}{\percent} in parameters, \SI{90}{\percent} in memory, and \SI{63}{\percent} in latency, and still produces the lower cost for \SI{55}{\percent} of the buildings.
Overall, these findings demonstrate that \gls{kd} is a practical and reliable step for deploying \gls{dt}-based battery control in hardware-restricted residential energy management systems. Future work will investigate alternative distillation objectives, assess the influence of offline trajectory quality, and extend the approach to more complex residential control scenarios involving multiple flexible devices.

\begin{acks}
We acknowledge support by the KIT-Publication Fund of the Karlsruhe Institute of Technology and by the state of Baden-Württemberg through bwHPC. R. Mikut and V. Hagenmeyer are supported by the Helmholtz Association in the Program Energy System Design.
\end{acks}

\bibliographystyle{ACM-Reference-Format}
\bibliography{references}

\clearpage
\appendix

\section{Mixed-Integer Linear Programming}
\label{sec:milp}
To benchmark the learned control policies, we compute an optimal schedule using a mixed-integer programming (MIP) formulation of the battery energy storage scheduling problem. The MIP yields a globally optimal solution with respect to the considered cost objective and therefore provides a deterministic lower bound on the achievable electricity cost. However, this benchmark is purely theoretical, since the optimisation assumes perfect knowledge of the complete prosumption and price trajectories over the full horizon and thus cannot be implemented in real time.

Given the prosumption and the electricity price trajectory, the optimisation determines a cost-minimising battery schedule over $T$ time steps. At each time step $t$, the decision variable $A_t$ denotes the battery charging power ($A_t>0$) or discharging power ($A_t<0$), bounded by the device limits. The battery's \gls{soe} evolves according to

\begin{equation}
     \mathrm{SoE}_{t+1} = \mathrm{SoE}_{t} + A_t,
\end{equation}

with $\mathrm{SoE}_t$ constrained to remain within the admissible capacity range for all $t$.

To model the asymmetric pricing scheme, a binary variable $b_t$ indicates whether the net grid exchange $\mathrm{Prosumption}_t + A_t$ corresponds to import ($b_t = 1$) or export ($b_t = 0$). Using big-$M$ and indicator constraints, the effective price $z_t$ is selected as the time-varying import tariff for $b_t=1$ and as a constant feed-in tariff for $b_t=0$, thereby capturing the piecewise pricing rule.
The overall electricity cost is minimised as
\begin{equation}
    \min \sum_{t=1}^{T} \left(\mathrm{Prosumption}_t + A_t \right) z_t .
\end{equation}

We enforce $A_t \in [ -A^{\max}_{\mathrm{dis}},\, A^{\max}_{\mathrm{ch}} ]$, $\mathrm{SoE}_t \in [\mathrm{SoE}^{\min},\, \mathrm{SoE}^{\max}]$, $b_t \in \{0,1\}$, and $z_t$ is restricted to the feasible tariff range. The optimisation problem is solved using Gurobi \cite{gurobi}.

\section{Hyperparameter}
\label{sec:hyperparameter}
\autoref{tab:hyperparameters_dt_kd} summarises the most relevant hyperparameters used in the \gls{dt} and \gls{kd} experiments. For these, the forecast horizon matches $h$ in \autoref{eq:states_vector} and the context length parameter applies to the \gls{dt} architecture's input.
\begin{table}[t!]
\centering
\caption{Hyperparameters for Decision Transformer \& Knowledge Distillation.}
\label{tab:hyperparameters_dt_kd}
\begin{tabular}{@{}ll@{}}
\toprule
\textbf{Hyperparameter} & \textbf{Value} \\ \midrule
Batch size & 32 \\
Forecast Horizon & 24 \\
Context Length & 96 \\
Optimiser DT & Adam \\
Optimiser DT Learning Rate & 1e-4 \\
Optimiser DT Weight Decay & 1e-4 \\
Optimiser KD & Adam \\
Optimiser KD Learning Rate & 1e-4 \\
Early Stop Patience & 500 \\ \bottomrule
\end{tabular}
\end{table}

\section{Detailed Results}
\begin{figure}[t!]
    \centering
    \includegraphics[width=\linewidth]{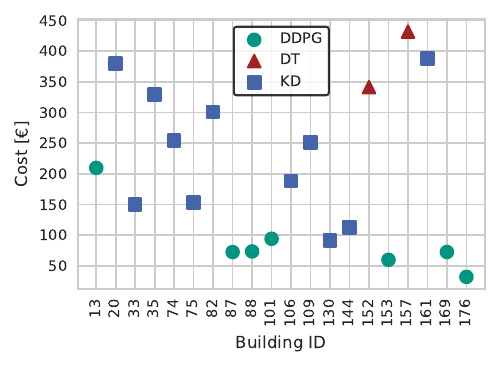}
    \caption{Best Performing Learning-Based Approach for Each of the 20 Buildings. DT Denotes the \textit{Medium} Decision Transformer, while KD Refers to Knowledge Distillation From \textit{Medium} Teacher to \textit{Tiny} Student.}
    \label{fig:comp_best_over_buildings_tiny}
\end{figure}
\autoref{fig:comp_best_over_buildings_tiny} summarises the best-performing learning-based control approaches evaluated across all 20 buildings. It compares the \textit{medium} \gls{dt} with its knowledge-distilled counterpart, where a \textit{tiny} student model is trained from the \textit{medium} DT teacher, and the \gls{ddpg} trajectory, highlighting the effectiveness of \gls{kd} under heavy resource constraints. \autoref{tab:dt_results} reports the detailed results of the three benchmarks, \gls{milp}, without battery, and \gls{ddpg}, in comparison with the different \gls{dt} model sizes. Furthermore, \autoref{tab:kd_results_small}, \autoref{tab:kd_results_medium}, and \autoref{tab:kd_results_large} each present the corresponding teacher model from \autoref{tab:dt_results} together with all three of its associated student models.
\begin{table*}[t!]
\centering
\caption{Cost in € Across All 20 Buildings for the Three Benchmarks and Five DT Variants, Averaged Over Three Random Seeds. The Final Row Reports the Mean Cost for Each Approach.}
\label{tab:dt_results}
\begin{tabular}{@{}lllllllll@{}}
\toprule
\textbf{Bld. ID} & \textbf{MILP} & \textbf{WO Battery} & \textbf{DDPG} & \textbf{DT Tiny} & \textbf{DT Mini} & \textbf{DT Small} & \textbf{DT Medium} & \textbf{DT Large} \\ \midrule
13 & 181.72 & 230.65 & 209.88 ± 4.8 & 210.79 ± 1.49 & 210.6 ± 1.62 & 210.68 ± 1.8 & 212.52 ± 0.66 & 210.29 ± 2.87 \\
20 & 357.44 & 392.74 & 384.01 ± 3.11 & 380.62 ± 1.22 & 381.22 ± 0.88 & 381.13 ± 1.0 & 380.34 ± 1.5 & 379.88 ± 0.4 \\
33 & 132.88 & 167.11 & 152.47 ± 2.03 & 151.81 ± 1.33 & 151.43 ± 0.73 & 152.18 ± 0.63 & 151.25 ± 1.33 & 152.08 ± 1.54 \\
35 & 314.47 & 343.74 & 334.37 ± 2.85 & 330.8 ± 1.95 & 330.45 ± 1.53 & 330.34 ± 1.12 & 329.38 ± 2.11 & 329.33 ± 0.97 \\
74 & 205.15 & 289.89 & 256.58 ± 1.33 & 258.81 ± 4.53 & 255.26 ± 3.47 & 255.7 ± 3.01 & 254.9 ± 3.78 & 266.91 ± 6.73 \\
75 & 70.97 & 206.03 & 170.22 ± 8.77 & 160.28 ± 6.81 & 156.06 ± 6.77 & 157.26 ± 4.14 & 154.93 ± 6.09 & 169.38 ± 9.64 \\
82 & 264.21 & 319.58 & 306.73 ± 3.98 & 302.32 ± 0.32 & 303.41 ± 2.66 & 301.66 ± 2.02 & 303.78 ± 3.14 & 301.22 ± 1.36 \\
87 & 51.14 & 89.64 & 72.72 ± 1.26 & 77.61 ± 1.45 & 77.05 ± 2.72 & 77.46 ± 2.66 & 77.84 ± 1.36 & 77.08 ± 0.23 \\
88 & 52.38 & 83.24 & 73.7 ± 1.31 & 76.52 ± 1.31 & 77.05 ± 0.47 & 77.2 ± 1.58 & 77.45 ± 1.44 & 75.21 ± 1.39 \\
101 & 74.53 & 110.97 & 94.38 ± 1.09 & 97.05 ± 1.71 & 96.54 ± 1.76 & 97.05 ± 1.79 & 96.15 ± 0.63 & 95.59 ± 1.93 \\
106 & 172.28 & 197.8 & 191.0 ± 3.2 & 189.52 ± 0.88 & 188.71 ± 0.37 & 188.69 ± 0.5 & 188.92 ± 1.42 & 189.9 ± 1.03 \\
109 & 211.97 & 278.7 & 251.74 ± 2.35 & 253.76 ± 2.74 & 252.79 ± 1.73 & 253.86 ± 2.63 & 251.3 ± 2.08 & 254.25 ± 6.49 \\
130 & 76.33 & 99.25 & 95.25 ± 2.09 & 92.38 ± 0.55 & 91.85 ± 0.77 & 91.7 ± 0.41 & 91.95 ± 0.87 & 91.8 ± 0.72 \\
144 & 95.42 & 129.35 & 113.76 ± 0.66 & 115.23 ± 2.27 & 114.82 ± 0.85 & 114.84 ± 0.86 & 113.78 ± 0.92 & 113.81 ± 0.12 \\
152 & 314.82 & 365.03 & 349.19 ± 1.0 & 343.43 ± 1.52 & 342.85 ± 0.73 & 342.0 ± 0.77 & 341.48 ± 0.7 & 340.06 ± 1.75 \\
153 & 40.78 & 85.89 & 60.1 ± 0.63 & 67.16 ± 4.1 & 66.5 ± 2.89 & 67.42 ± 2.4 & 66.93 ± 2.01 & 65.7 ± 0.49 \\
157 & 315.57 & 482.42 & 435.7 ± 15.93 & 432.2 ± 8.61 & 428.13 ± 5.18 & 429.62 ± 3.19 & 431.84 ± 2.51 & 464.11 ± 34.23 \\
161 & 362.43 & 400.62 & 391.78 ± 5.66 & 389.18 ± 0.56 & 389.51 ± 0.72 & 389.56 ± 0.57 & 388.26 ± 1.38 & 387.72 ± 1.48 \\
169 & 50.97 & 92.58 & 72.81 ± 2.49 & 78.04 ± 3.68 & 77.1 ± 1.66 & 76.69 ± 1.01 & 75.79 ± 3.15 & 74.96 ± 1.1 \\
176 & 16.21 & 39.67 & 32.27 ± 2.92 & 36.22 ± 1.71 & 36.92 ± 2.82 & 37.64 ± 1.74 & 37.11 ± 2.79 & 38.23 ± 0.53 \\ \midrule
 & 168.08 & 220.25 & 202.43 ± 3.37 & 202.19 ± 2.44 & 201.41 ± 2.02 & 201.63 ± 1.69 & 201.3 ± 1.99 & 203.88 ± 3.75 \\ \bottomrule
\end{tabular}
\end{table*}

\begin{table*}[h]
\centering
\caption{Cost in € Across All 20 Buildings for the Small Teacher and the Corresponding Student Models, Averaged Over Three Random Seeds. The Final Row Reports the Mean Cost for Each Approach.}
\label{tab:kd_results_small}
\begin{tabular}{@{}lllll@{}}
\toprule
\textbf{\begin{tabular}[c]{@{}l@{}}Bld.\\ ID\end{tabular}} & \textbf{\begin{tabular}[c]{@{}l@{}}Small\\ Teacher\end{tabular}} & \textbf{\begin{tabular}[c]{@{}l@{}}Tiny\\ Student\end{tabular}} & \textbf{\begin{tabular}[c]{@{}l@{}}Mini\\ Student\end{tabular}} & \textbf{\begin{tabular}[c]{@{}l@{}}Small\\ Student\end{tabular}} \\ \midrule
13 & 210.68 ± 1.8 & 210.51 ± 1.28 & 210.35 ± 2.49 & 209.93 ± 1.87 \\
20 & 381.13 ± 1.0 & 380.65 ± 1.0 & 380.21 ± 0.42 & 380.19 ± 0.16 \\
33 & 152.18 ± 0.63 & 151.84 ± 1.01 & 151.22 ± 0.4 & 151.36 ± 0.8 \\
35 & 330.34 ± 1.12 & 330.25 ± 0.71 & 329.5 ± 0.66 & 329.49 ± 0.58 \\
74 & 255.7 ± 3.01 & 254.8 ± 1.25 & 253.84 ± 2.09 & 253.56 ± 1.83 \\
75 & 157.26 ± 4.14 & 159.02 ± 6.36 & 155.81 ± 6.13 & 155.21 ± 5.57 \\
82 & 301.66 ± 2.02 & 302.03 ± 1.52 & 302.15 ± 0.84 & 302.83 ± 1.37 \\
87 & 77.46 ± 2.66 & 77.26 ± 1.48 & 76.78 ± 1.21 & 77.07 ± 1.18 \\
88 & 77.2 ± 1.58 & 76.16 ± 1.26 & 76.24 ± 1.62 & 76.71 ± 1.71 \\
101 & 97.05 ± 1.79 & 96.89 ± 0.89 & 96.3 ± 0.66 & 95.76 ± 1.06 \\
106 & 188.69 ± 0.5 & 187.94 ± 0.2 & 187.62 ± 0.51 & 187.69 ± 0.43 \\
109 & 253.86 ± 2.63 & 253.96 ± 3.44 & 250.87 ± 0.73 & 251.4 ± 0.83 \\
130 & 91.7 ± 0.41 & 91.15 ± 0.2 & 91.01 ± 0.34 & 91.32 ± 0.52 \\
144 & 114.84 ± 0.86 & 114.57 ± 0.57 & 113.82 ± 0.43 & 114.55 ± 0.35 \\
152 & 342.0 ± 0.77 & 342.16 ± 0.59 & 341.54 ± 0.32 & 341.32 ± 1.17 \\
153 & 67.42 ± 2.4 & 67.92 ± 2.14 & 67.31 ± 0.33 & 69.24 ± 1.02 \\
157 & 429.62 ± 3.19 & 434.72 ± 4.13 & 432.08 ± 5.11 & 429.86 ± 5.31 \\
161 & 389.56 ± 0.57 & 389.05 ± 0.97 & 388.87 ± 0.8 & 388.97 ± 0.82 \\
169 & 76.69 ± 1.01 & 77.62 ± 1.08 & 76.03 ± 0.95 & 77.08 ± 0.71 \\
176 & 37.64 ± 1.74 & 37.06 ± 1.88 & 37.42 ± 1.07 & 37.76 ± 1.36 \\ \midrule
 & 201.63 ± 1.69 & 201.78 ± 1.6 & 200.95 ± 1.36 & 201.06 ± 1.43 \\ \bottomrule
\end{tabular}
\end{table*}

\begin{table*}[h]
\centering
\caption{Cost in € Across All 20 Buildings for the Medium Teacher and the Corresponding Student Models, Averaged Over Three Random Seeds. The Final Row Reports the Mean Cost for Each Approach.}
\label{tab:kd_results_medium}
\begin{tabular}{@{}lllll@{}}
\toprule
\textbf{\begin{tabular}[c]{@{}l@{}}Bld.\\ ID\end{tabular}} & \textbf{\begin{tabular}[c]{@{}l@{}}Medium\\ Teacher\end{tabular}} & \textbf{\begin{tabular}[c]{@{}l@{}}Tiny\\ Student\end{tabular}} & \textbf{\begin{tabular}[c]{@{}l@{}}Mini\\ Student\end{tabular}} & \textbf{\begin{tabular}[c]{@{}l@{}}Small\\ Student\end{tabular}} \\ \midrule
13 & 212.52 ± 0.66 & 210.29 ± 1.17 & 211.02 ± 1.08 & 211.42 ± 0.45 \\
20 & 380.34 ± 1.5 & 379.61 ± 0.04 & 379.25 ± 0.46 & 379.2 ± 0.11 \\
33 & 151.25 ± 1.33 & 150.45 ± 0.44 & 150.56 ± 0.22 & 150.61 ± 0.54 \\
35 & 329.38 ± 2.11 & 329.07 ± 1.43 & 328.85 ± 1.46 & 328.6 ± 1.32 \\
74 & 254.9 ± 3.78 & 254.12 ± 2.44 & 254.99 ± 4.33 & 253.49 ± 2.55 \\
75 & 154.93 ± 6.09 & 153.1 ± 5.81 & 152.8 ± 5.89 & 152.19 ± 6.45 \\
82 & 303.78 ± 3.14 & 301.24 ± 1.16 & 302.55 ± 1.84 & 300.85 ± 0.84 \\
87 & 77.84 ± 1.36 & 76.83 ± 0.96 & 78.13 ± 1.38 & 77.22 ± 2.39 \\
88 & 77.45 ± 1.44 & 74.86 ± 1.86 & 74.18 ± 1.41 & 75.04 ± 2.34 \\
101 & 96.15 ± 0.63 & 96.72 ± 2.29 & 95.84 ± 0.86 & 96.02 ± 0.28 \\
106 & 188.92 ± 1.42 & 188.71 ± 0.86 & 188.43 ± 0.32 & 188.3 ± 0.81 \\
109 & 251.3 ± 2.08 & 251.16 ± 1.29 & 253.16 ± 1.49 & 251.43 ± 0.88 \\
130 & 91.95 ± 0.87 & 91.23 ± 0.64 & 91.27 ± 0.36 & 91.54 ± 0.31 \\
144 & 113.78 ± 0.92 & 113.07 ± 0.65 & 112.99 ± 0.85 & 113.05 ± 0.83 \\
152 & 341.48 ± 0.7 & 342.5 ± 2.89 & 341.33 ± 0.57 & 340.3 ± 0.9 \\
153 & 66.93 ± 2.01 & 66.23 ± 1.54 & 66.44 ± 2.96 & 65.81 ± 3.85 \\
157 & 431.84 ± 2.51 & 432.89 ± 3.21 & 429.82 ± 1.31 & 429.89 ± 1.48 \\
161 & 388.26 ± 1.38 & 387.7 ± 0.49 & 388.0 ± 0.9 & 388.0 ± 1.0 \\
169 & 75.79 ± 3.15 & 76.25 ± 2.08 & 75.26 ± 1.71 & 75.22 ± 2.09 \\
176 & 37.11 ± 2.79 & 37.95 ± 1.75 & 37.44 ± 1.21 & 37.34 ± 2.52 \\ \midrule
 & 201.3 ± 1.99 & 200.7 ± 1.65 & 200.62 ± 1.53 & 200.28 ± 1.6 \\ \bottomrule
\end{tabular}
\end{table*}

\begin{table*}[h]
\centering
\caption{Cost in € Across All 20 Buildings for the Large Teacher and the Corresponding Student Models, Averaged Over Three Random Seeds. The Final Row Reports the Mean Cost for Each Approach.}
\label{tab:kd_results_large}
\begin{tabular}{@{}lllll@{}}
\toprule
\textbf{\begin{tabular}[c]{@{}l@{}}Bld.\\ ID\end{tabular}} & \textbf{\begin{tabular}[c]{@{}l@{}}Large\\ Teacher\end{tabular}} & \textbf{\begin{tabular}[c]{@{}l@{}}Tiny\\ Student\end{tabular}} & \textbf{\begin{tabular}[c]{@{}l@{}}Mini\\ Student\end{tabular}} & \textbf{\begin{tabular}[c]{@{}l@{}}Small\\ Student\end{tabular}} \\ \midrule
13 & 210.29 ± 2.87 & 209.79 ± 0.4 & 211.61 ± 0.41 & 210.36 ± 2.16 \\
20 & 379.88 ± 0.4 & 379.91 ± 0.96 & 379.52 ± 0.23 & 379.4 ± 0.36 \\
33 & 152.08 ± 1.54 & 150.59 ± 1.01 & 150.65 ± 0.64 & 150.42 ± 0.63 \\
35 & 329.33 ± 0.97 & 329.47 ± 0.49 & 329.37 ± 0.66 & 329.22 ± 0.71 \\
74 & 266.91 ± 6.73 & 262.12 ± 8.92 & 262.58 ± 9.37 & 262.96 ± 8.49 \\
75 & 169.38 ± 9.64 & 164.06 ± 11.6 & 162.35 ± 14.59 & 162.47 ± 15.79 \\
82 & 301.22 ± 1.36 & 301.5 ± 0.9 & 300.19 ± 1.97 & 300.19 ± 2.24 \\
87 & 77.08 ± 0.23 & 76.7 ± 1.17 & 77.43 ± 1.62 & 77.38 ± 1.13 \\
88 & 75.21 ± 1.39 & 77.32 ± 4.05 & 77.27 ± 3.49 & 75.53 ± 1.82 \\
101 & 95.59 ± 1.93 & 95.19 ± 0.11 & 96.23 ± 0.77 & 95.82 ± 0.19 \\
106 & 189.9 ± 1.03 & 189.83 ± 0.62 & 189.09 ± 0.71 & 189.35 ± 0.75 \\
109 & 254.25 ± 6.49 & 255.13 ± 7.16 & 254.52 ± 7.23 & 253.93 ± 7.47 \\
130 & 91.8 ± 0.72 & 92.06 ± 0.1 & 91.85 ± 0.63 & 91.91 ± 0.55 \\
144 & 113.81 ± 0.12 & 113.42 ± 0.64 & 113.25 ± 0.1 & 113.14 ± 0.3 \\
152 & 340.06 ± 1.75 & 340.4 ± 1.43 & 340.19 ± 1.17 & 339.47 ± 1.34 \\
153 & 65.7 ± 0.49 & 66.01 ± 0.84 & 67.95 ± 2.42 & 66.04 ± 0.38 \\
157 & 464.11 ± 34.23 & 459.83 ± 31.21 & 450.36 ± 29.95 & 462.52 ± 33.31 \\
161 & 387.72 ± 1.48 & 388.13 ± 1.93 & 388.23 ± 1.93 & 388.1 ± 2.36 \\
169 & 74.96 ± 1.1 & 75.81 ± 1.03 & 75.21 ± 1.13 & 75.39 ± 0.8 \\
176 & 38.23 ± 0.53 & 37.1 ± 1.57 & 37.57 ± 1.41 & 38.14 ± 0.99 \\ \midrule
 & 203.88 ± 3.75 & 203.22 ± 3.81 & 202.77 ± 4.02 & 203.09 ± 4.09 \\ \bottomrule
\end{tabular}
\end{table*}

\end{document}